\def\year{2020}\relax
\documentclass[letterpaper]{article} 
\usepackage{aaai20}  
\usepackage{times}  
\usepackage{helvet} 
\usepackage{courier}  
\usepackage[hyphens]{url}  
\usepackage{graphicx} 
\usepackage{graphicx}  
\usepackage{mathtools}
\usepackage{amssymb}
\usepackage{amsmath}
\usepackage{booktabs}
\usepackage{subcaption}
\usepackage[ruled,vlined]{algorithm2e}
\usepackage{multirow}
\usepackage[frozencache]{minted}

\title{Approximate Latent Force Model Inference}
\author{Jacob D. Moss, Felix L. Opolka, Bianca Dumitrascu, Pietro Li\`{o}\\ 
University of Cambridge\\
Cambridge, United Kingdom\\
jm2311@cam.ac.uk 
}

\newcommand{\citet}[1]{\citeauthor{#1} \shortcite{#1}}
\newcommand{\citep}{\cite}
\newcommand{\KK}{\bd{K}_{*M}\bd{K}_{MM}^{-1}}
\newcommand{\alfi}{\textit{Alfi}}
\newcommand{\atcp}[1]{\tcp*[r]{\makebox[35mm]{#1\hfill}}}
\newcommand{\diff}{\mathrm{d}}
\newcommand{\E}[2][]{\mathbb{E}_{#1}\left[#2\right]}
\newcommand{\bd}[1]{\mathbf{#1}}

\begin{document}

\maketitle

\begin{abstract}
Physically-inspired latent force models offer an interpretable alternative to purely data driven tools for inference in dynamical systems. They carry the structure of differential equations and the flexibility of Gaussian processes, yielding interpretable parameters and dynamics-imposed latent functions. However, the existing inference techniques associated with these models rely on the exact computation of posterior kernel terms which are seldom available in analytical form. Most applications relevant to practitioners, such as Hill equations or diffusion equations, are hence intractable. In this paper, we overcome these computational problems by proposing a variational solution to a general class of non-linear and parabolic partial differential equation latent force models. Further, we show that a neural operator approach can scale our model to thousands of instances, enabling fast, distributed computation. We demonstrate the efficacy and flexibility of our framework by achieving competitive performance on several tasks where the kernels are of varying degrees of tractability.
\end{abstract}

\section{Introduction}

A vast variety of phenomena in the sciences, from systems of interacting particles to the aerodynamics of aircraft, are represented by differential equations. This has led to a large body of work tackling these problems using numerical methods, such as the Runge-Kutta methods and finite element analysis. Leveraging these developments, machine learning approaches such as neural ordinary differential equations (ODEs) \citep{chen2018neural} and non-parametric ODEs \citep{heinonen2018learning} have become popular for modelling data with black-boxes. The main difference between these methods lies in how the equations are represented (neural networks and Gaussian processes, respectively). However, the lack of an explicit equation means that these methods lose intepretability and the ability to infer latent terms. 

Latent Force Models (LFMs) \citep{lawrence2007modelling,alvarez2009latent} are a hybrid paradigm combining the structure and extrapolation of explicit models with the flexibility of Gaussian processes. This fusion results in a ``loose'' differential equation model which has been demonstrated to be a flexible approach to problems in computational biology \citep{gao2008gaussian,lopez2019physically,moss2020gene} and stochastic control \citep{sarkka2018gaussian,cheng2020patient}. However, they require a sometimes intractable integration of the kernel function, severely limiting their usefulness.

\begin{figure}[t]
    \centering
    \includegraphics[width=\columnwidth]{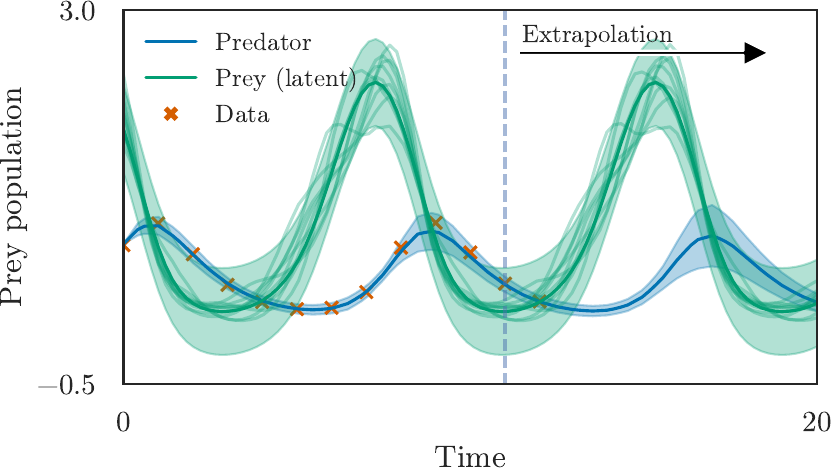}
    \caption{Exploiting known Lotka-Volterra population dynamics to obtain trajectories from data pertaining only to the predator. Conditioning the dynamics on data enables the model to extrapolate in time. Full experiment in Section \ref{sec: lotka}.}
    \label{fig:my_label}
\end{figure}
To address these drawbacks, we introduce \alfi{}, a flexible, efficient, variational framework for parameter estimation in ordinary and partial differential equation (PDE)-based LFMs. The main contributions of \alfi{} are as follows:



$\bullet\:$ \textbf{Efficient computation.} \alfi{} implements a gradient-matching pre-estimation procedure followed by fine-tuning with the full solution. It performs quadratic-time inference in LFMs in the number of inducing points, and is compatible with GPyTorch \citep{gardner2018gpytorch} modules. 

$\bullet\:$ \textbf{Scalability.} \alfi{} scales to large datasets by formulating a neural operator \citep{li2020fourier} architecture which learns a data-driven mapping between ODE/PDE solutions and latent forces. 
    
$\bullet\:$ \textbf{Flexible spatio-temporal analysis.} \alfi{} is able to infer latent forces from partially known dynamics in temporal and spatio-temporal settings without kernel derivations.

\section{Background} 

\textbf{Gaussian processes} are collections of random variables where each finite subset follows a multivariate Gaussian distribution as specified by the mean and covariance function. Inference takes the form of the posterior predictive distribution $p(\mathbf{y^*|y,x,x^*}) = \mathcal{N}(\mu_c, \Sigma_c)$ where $\mu_c = K_{\bf{x^*,x}}(K_{\mathbf{x,x}}+\sigma^2I)^{-1}\mathbf{x}$ and $\Sigma_c = K_{\bf{x,x}}-K_{\bf{x,x^*}}K_{\bf{x^*,x^*}}^{-1}K_{\bf{x^*,x}}$ from standard Gaussian identities.

\textbf{Latent Force Models} (LFMs) are differential equation models which leverage explicit dynamics to infer latent forcing terms \citep{lawrence2007modelling,alvarez2009latent}. The differential equation, $h$, parameterised by $\bd{\Theta}$, enforces a structural relationship between $P$ outputs, $\bd{y}(x)\in\mathbb{R}^P$, $N$ inputs, $\bd{x}\in\mathbb{R}^N$, and $L$ unobserved latent forces, $\bd{f}(x)\in\mathbb{R}^L$. GP priors are assigned to the latent forces, $f_i$: \(f_i\sim\mathcal{GP}(\textbf{0}, \kappa_i(x, x'))\), and these forces can be mixed by some response function $G(\bd{f})$.
\begin{equation}
    \overbrace{\mathcal{D}\,\bd{y}(\bd{x})}^{\textrm{differential}} = \overbrace{h\big(\bd{y}, \bd{x};\bd{\Theta}, G(f_1(\bd{x}),...,f_L(\bd{x}))\big)}^{\textrm{differential equation}}, \label{eq: lfm-general}
\end{equation}
where $\mathcal{D}$ is some differential operator, for example an n$^\text{th}$ order derivative for ODEs or partial derivatives for PDEs. 

An analytical expression for the covariance between outputs, $\kappa_{\bd{y},\bd{y}'}(t, t')$, is possible under the necessary condition that $G$ is a linear operator, meaning $G$ does not correlate the latent forces. In these cases, maximum marginal likelihood yields the differential equation parameters and inference can be carried out with standard posterior GP identities (see \citet{gaussianprocesses}). Larger datasets with non-linear latent forces lead to computational challenges.

\textbf{Variational Bayes} methods are used to approximate intractable integrals, such as when the prior, $p(\theta)$, is non-conjugate and a marginal likelihood, $p(\bd{x})= \int_\theta p(\bd{x}|\theta)p(\theta)\diff\theta$, is sought. 

\textbf{Partial differential equations} relate one or more state variables to their derivatives with respect to independent variables such as time and space. Some famous examples include the Navier-Stokes equations governing viscous fluids and the Schr\"{o}dinger equation for modelling the quantum wave-function. Many PDEs are not analytically solvable, leading to numerical approaches, such as the popular \textbf{Finite Element Method} (FEM). In FEM, the PDE is converted to a system of equations by finding its \textit{weak form} \citep{evans2010partial}. The PDE is then solved on a discretised spatial domain (mesh).

\textbf{Neural operators} are neural network representations of operators \citep{li2020fourier}, aiming to find a mapping between function spaces. They have been used to model the solution operator of PDEs, with benefits over traditional solvers such as mesh-invariance and are $\sim 1,000\times$ faster once trained. In this paper we use the Fourier variant, where pointwise transformations are learnt in the Fourier domain, thereby learning a global convolution in the physical domain. They require a large dataset of input-output samples.
\begin{figure*}[t]
    \centering
    \includegraphics[width=\textwidth]{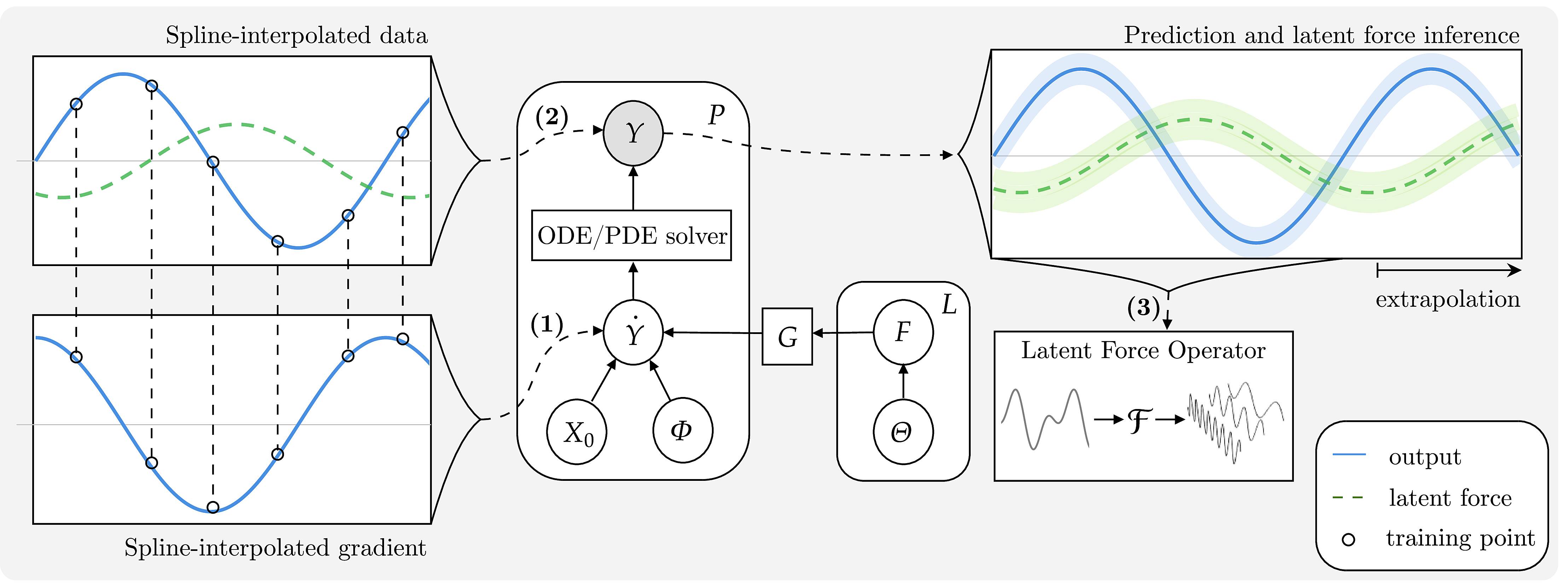}
    \caption{Computational graphical model for \alfi{}. \textbf{Step (1)} involves pre-estimating the parameters and latent forces of the differential equation in Eq. \ref{eq: lfm-general} by gradient-matching with the spline-interpolated derivative. \textbf{Step (2)} involves the more expensive forward solution of the differential equation and backpropagation through the ODE or PDE solver, for example using the adjoint equations, for fine-tuning the parameters. \textbf{Step (3)} optionally trains a Latent Force Operator, a neural network operating in the Fourier domain learning a mapping between solution and latent forces.}
    \label{fig:gm}
\end{figure*}

\newpage
\section{Approximate Latent Force Models}
In this section we propose
\alfi{}, an \textbf{a}pproximate \textbf{l}atent \textbf{f}orce \textbf{i}nference framework which significantly expands the class of models that can benefit from latent force interpretability. The cornerstone of \alfi{} is \textbf{flexibility}: it supports any differentiable kernel, and is agnostic to the functional form of the response to latent forces, $G$ \footnote{Indeed, $G$ could even take the form of a black-box MLP, however this results in loss of identifiability.}. For example, a neuroscientist analysing the behaviour of spike potentials may use a periodic kernel and a saturating response to the recovery variable \citep{fitzhugh1961impulses}. Such saturation functions are also commonly used in biophysical problems, such as the Hill equation for modelling reaction kinetics:
\begin{equation}
    G(\bd{f}(t);\textbf{w}_j) = \frac{\prod_i f_i(t)^{w_{ji}}}{\prod_i f_i(t)^{w_{ji}} + e^{-w_{j0}}} \label{eq:hill}
\end{equation}
where $w_{ji}$ are interaction weights. This non-linear operator thus incorporates cooperation/competition between the latent forces $f_i$, and has been shown to be a useful model for inspecting gene transcription networks \citep{titsias2012identifying}. However, such responses render the solution intractable. 

\subsection{Variational Objective}
A key motivation for this work is to avoid the onerous derivations of transformed kernels, which often have no closed form. To achieve this, \alfi{} employs well-studied variational inference techniques. Illustrated by the graphical model shown in Fig. \ref{fig:gm} are \alfi{}'s three main components, which are made possible through a variational objective.

Let $\bd{\hat Y}\in\mathbb{R}^{N\times P}$ represent the forward solution of an arbitrary differential equation at $N$ points, given parameters $\boldsymbol{\Phi}$, an arbitrary differentiable transformation ($G$) of latent forces, and input $\bd{X}\in\mathbb{R}^{N\times D}$. The input dimensions are $D=1$ (e.g. time) for ODEs, or $D>1$ (e.g. time and space) for PDEs. Our formulation extends to arbitrary output dimensions, $P$. Let $\bd{F}=\big[\bd{f}_i^\top\big]^L_i$ be our set of independent latent forces where $\bd{f}_i\sim \mathcal{GP}(0, \kappa_{\bd{\Theta}}(\bd{x},\bd{x}'))$ with kernel hyperparameters $\Theta$. We assume a Gaussian likelihood, $p(\bd{Y}|\bd{F},\boldsymbol{\Phi})=\mathcal{N}(\bd{Y}|\bd{\hat Y}, \sigma^2)$, and seek to optimise (hyper)parameters by marginalising over the latent forces:
\begin{align}
p(\bd{Y|\Phi,\Theta}) &= \int \overbrace{p(\bd{Y|F,\Phi})}^{\text{likelihood}}\prod_i^L\overbrace{p(\bd{f_i|X,\Theta})}^{\text{GP}}\diff \bd{f}.\label{eq:marg}
\end{align}
This marginal is intractable when $G$ is non-linear, so we derive an expectation lower-bound (ELBO) for the marginal likelihood using a variational approximation. We add $M$ inducing points, $\bd{u}_i\in\mathbb{R}^{M\times D}$, for each latent force: $\bd{U}=\big[\bd{u}_i^\top\big]_i^L$. Due to the conditional independence across the latent forces given any mixing parameters (e.g. Eq. \ref{eq:hill}), a factorised variational distribution is used: $q(\bd{u}) =\prod_i^L\mathcal{N}(\bd{m}_i, \bd{C}_i)$ with variational parameters $\bd{m}_i\in\mathbb{R}^M$ and $\bd{C}_i\in\mathbb{R}^{M\times M}$. This yields the ELBO we seek to to optimise:
\begin{equation}
\log p(\bd{Y|\Phi,\Theta}) = \E[q(\bd{F})]{\log p(\bd{y|F,\boldsymbol{\Phi}})} - \mathcal{KL}(q(\bd{U})||p(\bd{U})), \label{eq:expectation}
\end{equation}
where $q(\bd{F}) = \int p(\bd{F|U})q(\bd{U})\diff\bd{U}$ and we approximate the expected log-likelihood in Eq. \ref{eq:expectation} with Monte Carlo sampling. The full derivation for the ELBO is in Appendix \ref{app:elbo}. The variational posterior is $p(\bd{f}_i^*|\bd{u}_i) = \int p(\bd{f}_u^*|\bd{u}_i)q(\bd{u}_i) \diff \bd{u}_i,\label{eq:f|u} = \mathcal{N}(\bd{f}_i^*|\bd{m}_i^*, \bd{S}_i^*)$
where, as in \citet{hensman2015scalable}:
\begin{equation}
\begin{split}
    &\bd{m}^*_i = \KK\bd{m}_i\\
    \bd{S}^*_i = \bd{K}_{**} + &\KK(\bd{S}_i-\bd{K}_{MM})\bd{K}_{MM}^{-1}\bd{K}_{M*}
\end{split}
\end{equation}
\subsection{Inference Scheme}\label{sec:method}
We optimise the variational objective (Eq. \ref{eq:expectation}) as follows. At each epoch, a batch of the latent forces is sampled from the variational posterior using the reparameterisation trick, enabling pathwise gradient estimation. Parameters are pre-estimated in the first few epochs, followed by fine-tuning with the forward solution. Finally, a dataset can be generated from the LFM in order to train a Latent Force Operator. With $e$ differential equation evaluations, the computational complexity is $O(NM^2+e)$. Pseudocode along with sample code snippets are in Appendix \ref{app: scheme} and \ref{app: code}. We now detail these steps.

\textbf{Step (1): Parameter pre-estimation by gradient matching}

The forward solution of an ODE or PDE is the most computationally expensive component of our algorithm. As such, we develop a pre-estimation step to initialise the variational \textbf{and} differential equation parameters to sensible values. This step consists of a gradient-matching algorithm whereby the differential equation function is evaluated, with no calls to the solver, at all training points and compared with the interpolated gradient of the data at those points. We chose natural cubic splines for this purpose since they are linear in the number of points; more expensive alternative interpolations are discussed in Appendix \ref{app: scheme}.

\textbf{Step (2): Forward solution fine-tuning}
 
This step involves executing an appropriate differential equation solver to generate output samples from the LFM, with which we take the mean and variance under the Gaussian likelihood assumption.  

\textbf{For ODEs:} We calculate the forward solution using an ODE solver (including memory-efficient adjoint methods). Our method is solver-agnostic, although the choice affects when the latent forces are sampled. If using a fixed step-size solver, we can pre-compute the GP outputs at all the timepoints the solver will use. With adaptive step-sizes, we sample one point from the GP at each time-step.

\textbf{For PDEs:} Implementing PDE-based LFMs is more involved since they are typically much harder to solve than in the ODE case, and the forward solution is much more costly. The pre-estimation step is, therefore, crucial. Since cubic splines are used, only the 1st and 2nd derivatives are continuous, so if the model has higher order derivatives, our suggestion is to use a higher order spline or simply ignore those higher-order derivatives---only an approximation is desired in this step. The GP component uses an anisotropic kernel where each dimension gets a separate lengthscale hyperprameter. This is important as the latent force is likely to vary across time and space at different scales. 

In \alfi{}, we solve the PDE using FEM, which first requires converting it into its weak form. As an example use-case, we take reaction-diffusion equations, a ubiquitous second-order parabolic PDE used in the natural sciences:
\begin{equation}
    \frac{\partial u(x,t)}{\partial t} = \overbrace{\vphantom{\frac{x^2}{x^2}} R(x, t)}^{\text{reaction}} +  \overbrace{D\frac{\partial^2u(x,t)}{\partial x^2}}^{\text{diffusion}},
    \label{eq: reaction-diffusion}
\end{equation}
where $D$ is the diffusion constant(s). In this case, the differential operator is $\mathcal{D}=\partial/\partial t$ corresponding to the syntax in Eq. \ref{eq: lfm-general}. We go through the weak form for such equations in Section \ref{sec: reactdiff}. While these second-order PDE LFMs can have an explicit posterior, such as \citet{lopez2019physically}, they typically require complicated derivations involving Green's function. FEM is a much more flexible approach, enabling the solution of high-order PDEs across complex surfaces without requiring a new derivation on each iteration of the model. Since FEM is ubiquitous in engineering, this decision greatly increases the use-case coverage \citep{okereke2018finite}. 

\textbf{On weak form.} It is generally easier to obtain the weak (variational) form of a PDE than solving them analytically. As we will show in Section \ref{sec: reactdiff} (Eq. \ref{eq: weakform}), this form also neatly enables the incorporation of our latent forces. The reaction-diffusion systems we cover in this paper have provably unique weak solutions \citep{evans2010partial,croix2018bayesian}). 

\textbf{User-defined PDEs.} Solving PDEs using FEM first involves discretising the spatial domain by constructing a mesh made up of the \textit{elements} from which the method gets its name. This is the fundamental limitation of this class of PDE solvers. For 1D meshes, \alfi{} automatically constructs a mesh incorporating all vertices present in the dataset. Complicated meshes in XML format can be imported into \alfi{}. For example, the designer of a car might use a 3D model of the car to model forces during a crash. Once a mesh is selected, the weak form of the chosen PDE can be written in the Unified Form Language (UFL) \citep{alnaes2014unified}, a domain-specific language for writing weak forms. \alfi{} uses the \textit{FEniCS} package \citep{LoggMardalEtAl2012a,AlnaesBlechta2015a} for solving the variational problem as written in UFL.

\textbf{Gradients through the PDE solver.} Similarly with the ODE case, we need to propagate gradients through the finite element computation. Since PDEs in \alfi{} are written in their weak forms, the gradients can be determined with reverse-mode algorithmic differentiation using the adjoint equations. An implementation of the finite element method was provided by FEniCS, and adjoint gradients were computed using the \textit{dolfin-adjoint} package \citep{mitusch2019dolfin}, which automatically derives adjoints by constructing a computation graph and using the chain rule. A small wrapper \citep{torchfenics} can be used to propagate the gradients to PyTorch \textit{autograd} \citep{pytorch}.

\textbf{Step (3): Inductive learning} 

So far, we have dealt with a single instance of a physical system (for example, a reaction-diffusion system with a set of latent forces and an instantiation of parameters). For every new instance, a new model needs to be trained. This is acceptable for some low-dimensional ODE models, but for higher dimensions and PDEs this is computationally infeasible. We can scale up latent force inference in this setting by leveraging the generative nature of our variational LFM. Once a differential equation is chosen, we can generate a dataset with which to train a neural operator. At prediction time, such an operator would be significantly faster than training a new model (up to 20,000 faster in our experiments--see Appendix \ref{app: time}). We refer to this model as the Latent Force Operator (LFO), following the Fourier Neural Operator architecture \citep{li2020fourier}. An LFO aims to learn a mapping between the PDE or ODE solution space with the latent force space. Moreover, there are many examples of cases where a federated learning paradigm is desirable, for example patient time-series data \citep{cheng2020patient}. The structure of the LFO lends itself to distributed deployment, whereby inference on patients not in the training data is still possible. 

\begin{figure}[t]
    \centering
    \includegraphics[width=0.5\columnwidth]{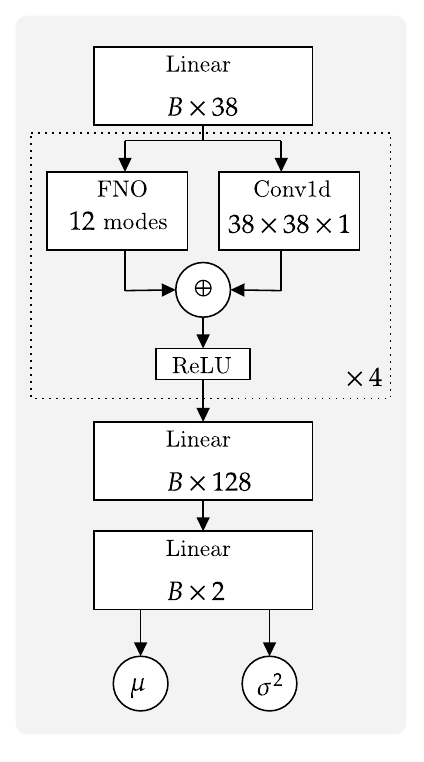}
    \caption{Latent Force Operator Architecture. $B$ is batch size.}
    \label{fig:lfo arch}
\end{figure}

As shown in Figure \ref{fig:lfo arch}, the architecture of an LFO consists of 4 Fourier Neural Operator (FNO) \citep{li2020fourier} layers and a 1D convolution capturing boundary conditions. The LFO outputs the mean and variance of a Gaussian and is trained by maximum likelihood.


\section{Experiments}
In order to test the performance and generality of \alfi{}, we apply our method to a number of tasks from different domains. Starting with a simulation of predator-prey dynamics, we demonstrate the kernel's impact on the ability to infer the prey dynamics without any knowledge of prey abundance. We then compare our approximation with exact solutions of models of mRNA production in response to transcription factors. In order to investigate PDE-based LFMs, we treat the spatio-temporal production of protein in fruit fly embryos with a reaction-diffusion equation. We finally demonstrate our Latent Force Operator can scale up inference to thousands of instances. 
\begin{figure}
    \centering
    \includegraphics[clip=true, width=\columnwidth]{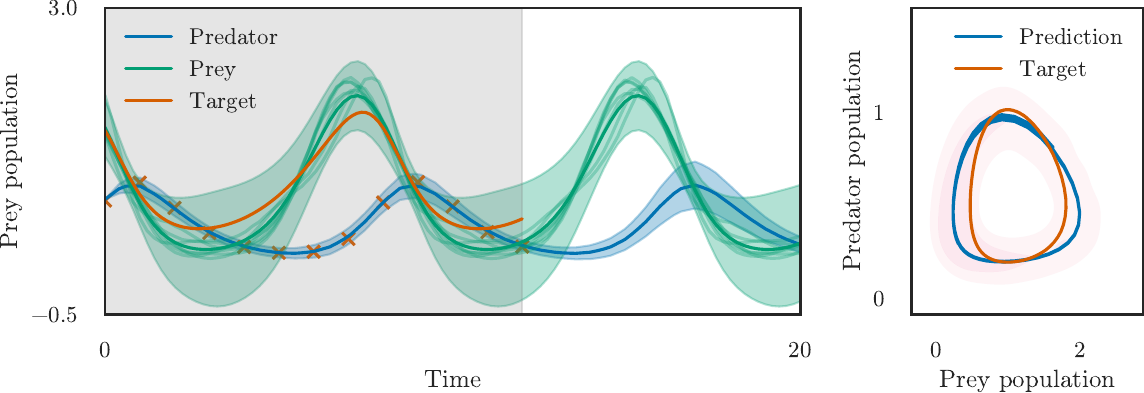}
    \caption{\alfi{} with a periodic kernel on the Lotka-Volterra task. Shaded area is the training range, orange crosses indicate the training datapoints. Blue and green lines are used for and predator and prey respectively, orange line for the \textbf{unseen} prey ground truth, and translucent lines for sample trajectories. The filled intervals show a $\pm 2\sigma$ range. \textit{Left:} the output dynamics with the training data as crosses along with the latent dynamics. \textit{Right:} the phase space ($u,v$) alongside the unseen ground truth. The shaded region was produced using kernel density estimation.}
    \label{fig: lv_main_plot}
\end{figure}

We use the $Q^2=1-\mathrm{SMSE}$ criterion, where $\mathrm{SMSE}$ is the mean-squared-error (MSE) normalised by the target variance, overcoming the sensitivity of MSE to the scale of the target values \citep{gaussianprocesses}. In addition, to give more insights into the more challenging spatio-temporal experiments, we also quote coverage interval accuracy deviation ($68\%-\operatorname{CA}_{\pm\sigma}$), which measures the proportion of targets within 1 standard deviation of the predictive distribution.
\begin{figure*}[t]
     \centering
     \begin{subfigure}[b]{0.49\textwidth}
    \centering
    \includegraphics[width=\textwidth]{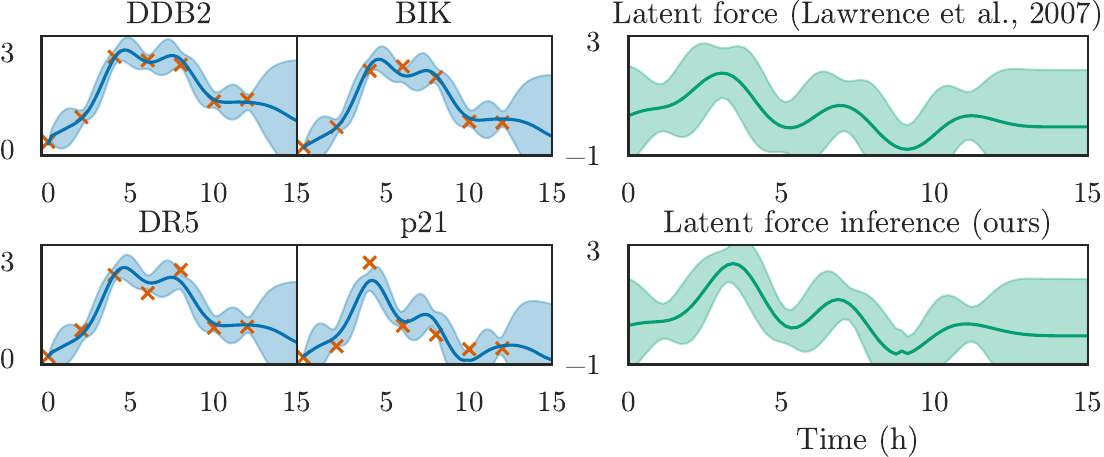}
    \caption{Variational approximation of the linear LFM from \citet{lawrence2007modelling} applied to the Brewer et al. (2006) microarray dataset. \textit{Left:} the output mRNA abundances of 4 genes: DDB2, BIK, DR5, p21, all of which are key in the DNA damage response pathway. \textit{Right top:} the latent force inferred by the exact LFM. \textit{Right bottom:} the latent force inferred by our variational LFM.}
    \label{fig:linear-p53}
     \end{subfigure}
     \hfill
     \begin{subfigure}[b]{0.49\textwidth}
    \centering
    \includegraphics[width=\textwidth]{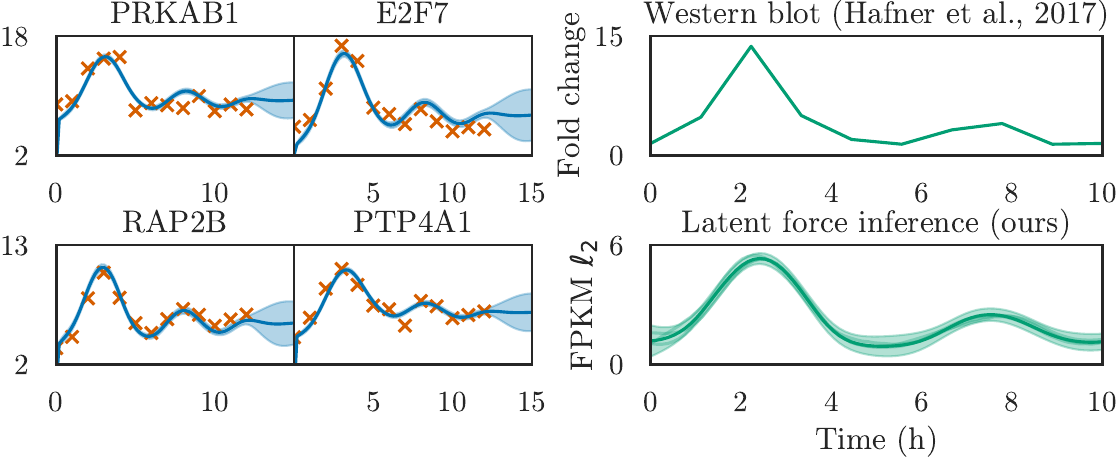}
    \caption{Variational approximation of a non-linear LFM applied to the \citet{hafner2017p53} RNA-seq dataset. \textit{Left:} the output abundances of 4 genes in $\ell_2$-normalised FPKM units. \textit{Right top:} the quantised protein fold change of a western blot in \citet{hafner2017p53}. \textit{Right bottom:} the inferred p53 mRNA abundance. Note that the units for the blot and inferred latent force are not comparable.}
    \label{fig:nonlinear-p53}
     \end{subfigure}
        \caption{Evaluating the performance of our variational LFM versus the exact solution and experimental data. In both, the latent force is the p53 protein.}
        \label{fig:transreg-lfm-comparisons}
\end{figure*}

\subsection{Lotka-Volterra}\label{sec: lotka}
The Lotka-Volterra equations are a system of non-linear first-order differential equations used to model predator-prey dynamics. The population of predators,  $v$, is related to the population of prey, $u$, with the following pair of equations:
\begin{equation}
    \frac{\diff u}{\diff t} = \alpha u - \beta uv, \qquad \frac{\diff v}{\diff t} = \delta uv - \gamma v,
\end{equation}
where $\alpha, \delta$ are growth rates; $\beta, \gamma$ are decay rates; and for clarity $u = u(t)$ and $v=v(t)$. 

\textbf{Latent Force Problem:} We assume that the population of prey is unknown and positive, as such we assign a Gaussian process prior to the function $u(t) \sim \mathcal{GP}(0, \kappa(t, t'))$ and a softplus response, $G(\bullet) = \log\left(1+e^\bullet\right)$, meaning the solution is intractable. We aim to recapitulate the correct latent force profile in addition to the growth and decay parameters ($\bd{\Phi} = \{\delta, \gamma\}$). While the the ubiquitous radial basis function (RBF), a universal smooth function approximator, would provide a good interpolation, a more sensible kernel would be periodic (the Lotka-Volterra solution is periodic):
\begin{equation}
    \kappa_{\text{p}}(t, t') = \exp\left(\frac{2\sin^2(\pi|t-t'|)/p}{\ell^2}\right),
\end{equation}
where $p$ is the period parameter, determining the interval of time after which the function repeats. As seen in Figure \ref{fig: lv_main_plot}, the periodic kernel enables realistic extrapolation beyond the training time range. As shown in Appendix \ref{app: lotka}, an RBF kernel yields inferior results. Moreover, a zero-mean function can lead to an extinction event outside the data range.

\subsection{Transcriptional Regulation}\label{sec: transreg}

In this experiment we test whether the approximations used in \alfi{} can produce results comparable to that which an exact solution produces. We take an ODE model model of the biological process of \textit{transcription} as an example. The time derivative of mRNA, $y_j(t)$, of gene $j$ is related to its latent regulating transcription factor protein(s) $f_i(t)$ \citep{barenco2006ranked}:
\begin{equation}
    \frac{\diff y_j(t)}{\diff t}=\overbrace{b_j}^{\mathclap{\textrm{basal rate}}}+s_j\overbrace{G(f_1(t),...,f_L(t);\textbf{w}_j,w_{j0})}^{\textrm{response}} - \overbrace{d_jy_j(t)}^{\textrm{decay term}} \label{eq:lawrence-ode}
\end{equation} 
The exact solution (see \citet{lawrence2007modelling}) is only tractable when the response function, $G$, is the identity. Moreover, we demonstrate that \alfi{} produces experimentally validated results when a non-linear response is used. For these purposes, we use two wet-lab datasets related to DNA damage response due to ionising radiation in radiotherapy. These are detailed in Appendix \ref{app: trans datasets}. 

\textbf{Latent Force Problem:} 
We assume that the protein abundance of a transcription factor (TF) is unknown, so we assign a Gaussian process prior with an RBF kernel to the function $f(t)\sim \mathcal{GP}(0, \kappa_{\text{r}}(t,t')$. This is a realistic scenario since there is not always a correlation between the TF mRNA level and its associated protein (see, e.g., \citet{gedeon2012delayed}).
\begin{table}[t]
  \caption{Results for transcriptional regulation. Lawrence = the linear LFM from Lawrence et al. (2007), VLFM = variational LFM, and LFO = latent force operator. * indicates our models. LFM models were trained for 700 epochs, and the LFO was trained for 50 epochs on a dataset of 2000 LFM solutions. The best $Q^2$ was selected and averaged over 15 generated test datasets. There is no output $Q^2$ for the LFO since it is not an output.}
  \label{tab: transreg results}
  \centering
  \begin{tabular}{p{11mm}llll}
    \toprule
    Model     & Output $Q^2$ $\uparrow$ & Latent $Q^2$ $\uparrow$ & Param. $\operatorname{MAE}$ $\downarrow$ \\
    \midrule
    Lawrence & $94.1 \pm 3.4$ & $92.6 \pm 4.4$ & $0.583 \pm 0.172$ \\
    VLFM* & $\bd{99.8 \pm 0.9}$ & $85.5 \pm 9.8$ & $0.668 \pm 0.182$ \\
    LFO* & -- & $95.6 \pm 3.0$ & $0.694 \pm 0.160$ \\
    \bottomrule
  \end{tabular}
\end{table}

For the first dataset, the same training setup is used as in \citet{lawrence2007modelling} consisting of 5 gene targets upregulated by the transcription factor protein p53. For the second, we take 22 gene targets and use a softplus response function instead, rendering the solution analytically intractable. This has the affect of constraining the latent force to be positive, a sensible assumption as negative abundance is nonsensical. The ODE parameters are denoted as $\bd{\Phi} = \{b_j, s_j, d_j\}_j^P$. 

\begin{figure}[t]
     \centering
    \includegraphics[width=\columnwidth]{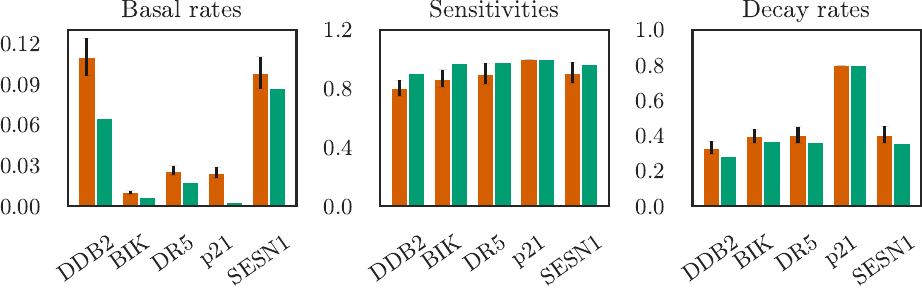}
    \caption{Kinetic parameters as estimated by our variational approximation compared to \citet{lawrence2007modelling}. Error bars are 1 standard deviation of 10 random model initialisations.}
    \label{fig:linear-kinetics}
\end{figure}

The results are shown in Figures \ref{fig:linear-p53} and \ref{fig:nonlinear-p53} for the linear and non-linear responses respectively. We compare our variational approximation to the analytical solution as derived in \citet{lawrence2007modelling} in terms of kinetic parameter inference in Figure \ref{fig:linear-kinetics}. In addition, we trained a Latent Force Operator on a dataset of 2,000 instances generated from this LFM with different values for $\bd{\Phi}$. Table \ref{tab: transreg results} shows the results comparing performance of our models compared with an analytical solution on a synthetically generated transcriptional regulation dataset. For the exact and variational LFMs, one gene's sensitivity parameter was fixed during training to maintain identifiability. For the neural operator, enforcing identifiability is an open problem. Time comparisons between the exact, variational, and neural operator variants are presented in Appendix \ref{app: time}.

\subsection{Reaction-Diffusion System}\label{sec: reactdiff}

We now demonstrate the performance of \alfi{} for dynamical systems modelled by partial differential equations, and compare the results to those found by tailored models. Reaction-diffusion equations are one of the most commonly used models of processes in the natural sciences, for example biology \citep{erneux1978turing,croix2018bayesian,lopez2019physically}) and chemistry \citep{lee1993pattern}. We use the reaction-diffusion equation in Eq. \ref{eq: reaction-diffusion}, setting the reaction term to:
\begin{equation}
    R(x,t) = Su(x,t) - \lambda y(x,t)
\end{equation}
where $S$ is the production rate of the driving mRNA (the latent force) and $\lambda$ is the decay rate, as in \citet{lopez2019physically}. Dirichlet boundary conditions are assigned such that $y(x=0,t) = 0$ and $y(x=l,t)=0$ where $0$ and $l$ are the spatial boundaries.

\textbf{Latent Force Problem:} We assign an anisotropic RBF kernel prior over the latent force $u$, giving us a lengthscale for each time and space dimension. The latent force represents the mRNA of the gap gene involved in Drosophila embryogenesis, while the output, $y$, is the protein concentration.

We start with deriving the weak form for this PDE by multiplying by a test function $v\in V$ and integrating the second derivatives by parts over the spatial domain, $\Omega$.
\begin{align}
\partial_t y \approx \frac{y^{n+1}-y^n}{\Delta t} &= Su - \lambda y +  D\nabla_x^2y, \\
y^{n+1} - \Delta t D\nabla^2 y^{n+1} &= y^n + \Delta t Su - \Delta t\lambda y^{n+1},\\
\begin{split}
    \int_\Omega yv + \Delta tD\nabla_x y\cdot \nabla_x v \:\diff \Vec{x} &=\\ \int_\Omega (y^n + &\Delta t Su - \Delta t\lambda y)v\: \diff \Vec{x},\label{eq: weakform}
\end{split}\\
a(y, v) &= L_{n+1}(v) 
\end{align}
where $a(y, v)$ is a bilinear form, $v$ is known as a test function belonging to some Banach space $V$, and function parameters have been omitted (i.e. $u = u(x, t)$). For completeness, the full derivation and transcription into UFL can be found in Appendix \ref{app:heat} and \ref{app: ufl}. 

\begin{figure}[t]
  \begin{center}
    \includegraphics[width=0.4\textwidth]{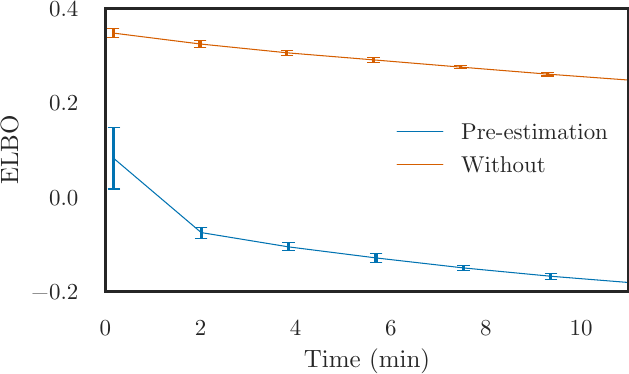}
  \end{center}
  \caption{Pre-estimation performance: gradient-matching prior to fine-tuning improves convergence.}
  \label{fig:preestimation perf}
\end{figure}

We compare our model to the GP-mRNA model in \citet{lopez2019physically} which uses an approximate solution using 20 terms of a Green's function, and also treats the mRNA as a latent force. The quantitative results are shown in Table \ref{tab: dros results}, and the full qualitative outputs are in Appendix \ref{app: dros}. We note that when using the parameter pre-estimation procedure, the convergence is significantly sped up and the ELBO starts at a lower initial value, as shown in Figure \ref{fig:preestimation perf}.

\begin{figure*}
    \centering
    \includegraphics[width=\textwidth]{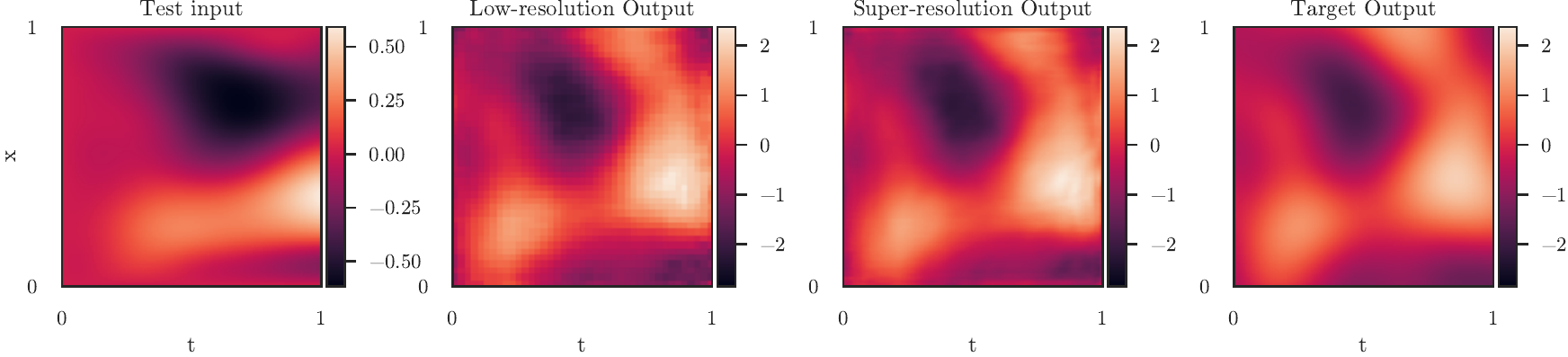}
    \caption{Neural operator for super-resolution of the synthetic reaction-diffusion dataset. \textit{From left:} The PDE solution input into the model; the low-resolution output (the training resolution, where input is subsampled); the super-resolution output (where the model has not been trained at this resolution); and the super-resolution ground truth. Graphics best seen on a computer screen.}
    \label{fig:toy neural operator}
\end{figure*}

In order to demonstrate Step (3) of \alfi{}, we generated a dataset of reaction-diffusion PDE solution-latent force pairs using Steps (1) and (2), using it to train a Latent Force Operator. The results are shown in Figure \ref{fig:toy neural operator}, demonstrating the zero-shot ability of the neural operator to predict at a higher mesh resolution.

\begin{table*}[h]
\caption{Results of \alfi{} versus the closest model in the literature, GP-mRNA \citep{lopez2019physically}. \alfi{} obtains a comparable performance in most cases, although the higher latent force inference error is an avenue for further research.}
\label{tab: dros results}
\centering
	\begin{tabular}{cccccc}
		\toprule
		&Trunk & \multicolumn{2}{c}{$Q^2 [\%]\uparrow$} & \multicolumn{2}{c}{$68-\operatorname{CA}_{\pm\sigma} [\%] $} \\
		Model & Gap & mRNA & Gap Protein & mRNA & Gap Protein \\
		& Gene & $\mu \pm \sigma$ & $\mu \pm \sigma$ & $\mu \pm \sigma$ & $\mu \pm \sigma$ \\
		\midrule
		& $kr$ & $\boldsymbol{90.5 \pm 2.0}$ & {$90.8 \pm 0.6$} & $-3.7 \pm 6.7$ & $21.8 \pm 4.0$ \\ 
		GP-mRNA & $kni$& {$81.1 \pm 2.5$} & {$88.7 \pm 0.8$} & $18.3 \pm 6.3$ & $35.3 \pm 5.3$  \\ 
		& $gt$ & $\bd{91.2 \pm 1.9}$ & $\bd{92.3 \pm 0.6}$ & $-0.7 \pm 3.3$ & $25.8 \pm 1.8$ \\ 
		\midrule
		& $kr$  & $82.4 \pm 1.2$ & $\bd{95.7 \pm 0.2}$ & ${3.3 \pm 3.5}$ & ${19.4 \pm 2.0}$ \\ 
		\alfi{}* & $kni$& $77.2 \pm 2.6$ & $86.9 \pm 2.9$ & $\bd{-1.5 \pm 1.6}$ & $\bd{12.4 \pm 9.1}$\\ 
		& $gt$ & $72.4 \pm 3.3$ & $86.5 \pm 1.2$ & $-3.9 \pm 4.4$ & $\bd{5.8 \pm 14.3}$ \\ 
		\bottomrule
	\end{tabular}
\end{table*}

\section{Related Work}

Machine learning methods for differential equations typically assume the equation is unknown, and follow either a GP or neural network approach. For example, differential GPs \citep{hegde2018deep}, generalise the discrete layers of a deep GP to a sequence of infinitesimal steps in which the input is warped through a vector field defined by an stochastic differential equation (SDE). However, the SDE acts only as a mechanism for constructing a model rather than as an actual physical model. A similar approach also treats the ODE as unknown using a Gaussian process and uses the full solution along with sensitivity equations to extract gradients \citep{heinonen2018learning}. This model, however, only uses the conditional mean of the Gaussian process, whereas \alfi{} also captures the covariance. The neural network methods, popularised by \citet{chen2018neural}, represent the ODE function as a neural network, where the weights are optimised by backpropagation through an ODE solver or via adjoint equations. Some recent extensions add stochasticity in the form of initial condition distributions \citep{norcliffe2021neural}. Unknown dynamics models lack the ability to infer latent forces, and therefore cannot be compared directly with LFMs. 

Latent force models (LFMs) are between the explicit and unknown differential equation paradigms; the equation depends on a set of latent forces which are inferred using the explicit dynamics. There are several other approaches for overcoming the intractable posterior in non-linear LFMs, namely the Laplace approximation \citep{lawrence2007modelling} and MCMC \citep{titsias2008efficient}. \citet{hartikainen2012state} show the connection between non-linear latent force models and non-linear SDEs, thereby enabling an approximation using Gaussian filtering and smoothing. Neural network approximations have also been attempted, where an inverse autoregressive flow \citep{kingma2016improving} is used to model the variational posterior \citep{ward2020black}.

\section{Discussion}\label{sec:conclusion}

In this paper, we present \alfi{}, a novel framework for approximate inference in hybrid mechanistic-Gaussian process models known as Latent Force Models. \alfi{} enables the implementation of linear, non-linear, ordinary, and partial LFMs without complex kernel derivations. We also introduce the Latent Force Operator which uses the generative nature of our approximate inference procedure in order to infer latent forces at a fraction of the cost of the standard LFM. If accurate parameter estimates are required, we recommend the LFM, however, if the number of instances is large, then the operator is more efficient. 

Unlike other approximate LFM methods, such as MCMC \citep{titsias2012identifying}, timepoints do not have to be specified \textit{a-priori}. The output of \alfi{} is a conditional Gaussian process, meaning the resolution, uniformity, and range of evaluation points is arbitrary. In addition, the error of the result is tunable using the tolerances in the equation solver. When compared with analytical LFMs, our method overcomes the most time-consuming component of implementing such models: the complex derivations of integrated kernel function and gradients, which are either intractable or involve complicated convolutions \citep{lawrence2007modelling}. \alfi{} also enables the use of any user-defined kernel.

Furthermore, non-parametric ODEs can be implemented with \alfi{} easily; the only modification required is to giving the state as input, rather than time. The ODE function will then sample from this Gaussian process in a similar manner to deep GPs \citep{damianou2013deep}. This results in a vector field in the state dimensions.

\textbf{Model assumptions:} With a Gaussian likelihood, we are assuming the latent forces are continuous. For counts data such as in the Lotka-Volterra and mRNA experiments, a discrete distribution such as the Poisson would be more appropriate; however, the chosen distribution must admit backpropagation. Moreover, the choice of equation and kernel function impresses a strong inductive bias which is responsible for the strong performance of LFMs. A periodic kernel in the Lotka-Volerra experiment, for example, enables the oscillatory steady state, without which, the prey (latent force) will revert to the GP mean as it gets further from the data, potentially leading to a stable extinction fixed point. 

\textbf{No free lunch:} the performance of parameter estimation differs between the exact solution and the approximation used by \alfi. Moreover, models like those presented in this paper estimate parameters over a collection of time-series which tend to lead to large distributions (so-called \textit{sloppiness} \citep{gutenkunst2007universally}). Such ``collective fits'' thus tend to leave parameters poorly constrained. A question for future work is to what extent are the LFM parameters interpretable, and whether the emphasis should instead be on predictive performance. Nonetheless, the mechanistic component enforces a structural relationship that enables the inference of latent forces, which is not possible otherwise. Since FEM is limited by the curse of dimensionality, so will \alfi{}'s performance on PDEs. Regardless, the severity is minimal since many physical problems are in 3D or 4D.
\newpage
\textbf{Broader impact:} \alfi{} greatly increases the class of models that can be tackled with latent force inference. We expect that this framework will therefore enable practitioners to construct more complex dynamical models. Moreover, the speed at which \alfi{} enables LFMs to be constructed makes iterating model design much easier.

\textbf{Ethical considerations:} \alfi{} is a framework for inferring latent forces in a dataset or data instance. As such, there could be sensitive applications (e.g. medical time-series) where blind faith in the inferred forces can lead to potentially detrimental effects on the patient. This is a consideration inherent in all models, however in this case the latent nature could be more pernicious. 

\textbf{Code availability:} the code for \alfi{} and the experiments is available here: https://github.com/mossjacob/alfi.
\newpage
\appendix

\section{Training Scheme} \label{app: scheme}
\textbf{Natural cubic splines:} Our pre-estimation procedure is similar to neural controlled differential equations \citet{kidger2020neural} only in the choice of spline. Our interpolant is not fed through a solver--it is used only for gradient estimation. There are many possible interpolation approaches to use instead of cubic splines. One popular method is using a Gaussian process \citep{calderhead2009accelerating,dondelinger2013ode,wenk2019fast} with a differentiable kernel (and mean) function. Gaussian processes with such kernels are closed under differentiation, and the covariance terms can be found as $\text{cov}(\dot y_j(t), \dot y_j(t'))=\frac{\partial^2 K(t,t')}{\partial t\partial t'}$. Cubic splines are linear in the number of points, in contrast to GPs which are cubic (or quadratic for non-naive implementation \citep{gardner2018gpytorch}).

\textbf{Natural gradient optimisation:} optimising the variational parameters using gradient descent in parameter space assumes Euclidean geometry. Variational parameters exist in distribution space, so we should use a more appropriate geometry within which we calculate distance, such as KL-divergence. Natural gradient optimisation takes steps in this space.

\begin{algorithm}[h]
\SetAlgoLined
\SetKwInOut{In}{Out}
\KwIn{$\bd{x,y}$}
 initialization\;
 \For{i $\leftarrow$ $1$ \KwTo $T$}{
  latent\_forces = variational\_posterior($\bd{x}$)\;
  \eIf{preestimation}{
   target $\leftarrow$ spline\_interpolate\_gradient($\bd{x, y}$) \atcp{target is gradient}
   pred $\leftarrow$ de\_func(initial\_conditions, latent\_forces)\;
   }{
   target $\leftarrow$ $\bd{y}$ \atcp{target is solutions}
   pred $\leftarrow$ solve(de\_func, initial\_conditions, latent\_forces)\;
  }
  loss $\leftarrow$ elbo(pred, target)
 }
 \caption{\alfi{} Scheme: Steps (1) and (2)}
\end{algorithm}

\section{ELBO Derivation}\label{app:elbo}
For completeness, we include here the derivation of the ELBO in Equation \ref{eq:expectation}.
\begin{align}
\log p(\bd{Y|\Theta,\Phi}) &=  \log\iint p(\bd{Y|F}) p(\bd{F|U})p(\bd{U})\diff \bd{U}\diff\bd{F},\\
= \log\iint &\frac{p(\bd{Y|F}) p(\bd{F|U})p(\bd{U})p(\bd{F|U})q(\bd{U})}{p(\bd{F|U})q(\bd{U})}\diff \bd{U}\diff\bd{F}
\end{align}
We now apply Jensen's inequality:
\begin{align}
&\geq \iint p(\bd{F|U})q(\bd{U})\log\frac{p(\bd{Y|F})p(\bd{U})}{q(\bd{U})}\diff \bd{U}\diff\bd{F},\\
&= \int q(\bd{U})\left[\int p(\bd{F|U})\log p(\bd{Y|F}) \diff\bd{F} + \log\frac{p(\bd{U})}{q(\bd{U})}\right]\diff \bd{U},\\
&= \E[q(\bd{U})]{\E[p(\bd{F|U})]{\log p(\bd{Y|F})}} -\mathcal{KL}(q(\bd{U})||p(\bd{U})) ,\\
&=\E[q(\bd{F})]{\log p(\bd{Y|F})} - \sum_i^L\mathcal{KL}(q(\bd{u}_i)||p(\bd{u}_i)),
\end{align}

\section{Heat Equation Weak Form Derivation}\label{app:heat}
We go through the complete instructive derivation for the heat equation's variational form below.
\begin{equation}
\partial_t y(x,t) = Su(x,t) - \lambda y(x,t) +  D\nabla_x^2y(x,t)
\end{equation}
The first step in FEM is to find the variational form. We henceforth omit function parameters (e.g. $u = u(x, t)$). First, we take the finite backwards difference as follows $\forall n=0,1,2,...$:
\begin{equation}
    \partial_t y^{n+1} \approx \frac{y^{n+1}-y^n}{\Delta t} = Su - \lambda y + D\nabla^2_x y
\end{equation}
We rearrange to find a bilinear form on the left-hand-side and a linear form on the right: 
\begin{equation}
    y^{n+1} + \Delta t\lambda y^{n+1} - \Delta t D\nabla^2 y^{n+1} = y^n + \Delta t Su
\end{equation}

For brevity we now drop the $n+1$ indices ($y=y^{n+1}$). The next step in FEM is to multiply both sides by a test function $v\in V$ and integrate over the spatial domain $\Omega$. The Dirichlet boundary conditions are defined such that $y(x=0,t) = 0$ and $y(x=l,t)=0$ where $0$ and $l$ are the spatial boundaries.
\begin{equation}
    \int_\Omega (yv + \Delta t\lambda yv - \Delta t D\nabla^2_x y v) \diff \Vec{x} = \int_\Omega(y^n+\Delta t Su )\,v\, \diff \Vec{x}
\end{equation}
We can integrate the second derivatives by parts:
\begin{equation}
    \int_\Omega \Delta tD\nabla^2_x yv\: \diff  \Vec{x}= D\underbrace{[\nabla_x y\cdot v]_0^l}_{0} - \int_\Omega \Delta tD\nabla_x y\cdot \nabla_x v \:\diff \Vec{x}
\end{equation}
where $[\nabla_x y\cdot v]_0^l$ vanished due to boundary conditions. This yields the variational (weak) form:
\begin{align}
    \int_\Omega yv + \Delta t\lambda yv + \Delta tD\nabla_x y\cdot \nabla_x v \:\diff x &= \int_\Omega (y^n + \Delta t Su)v\: \diff \Vec{x}\\
    a(y, v) &= L_{n+1}(v) 
\end{align}
This weak form can then be written in UFL almost verbatim.

\section{Code} \label{app: code}

\subsection{Differences between an ODE- or PDE-based LFM}
Both  \texttt{OrdinaryLFM()} (for ODE-based LFMs) and \texttt{PartialLFM()} (for PDE-based LFMs) inherit from the  \texttt{VariationalLFM()} class which provides methods for sampling from the variational GP. Depending on which class is used, the practitioner must implement the \texttt{ode\_func} or \texttt{pde\_func} function, which are used by the solvers to calculate the forward solution, and the pre-estimator (\texttt{PreEstimator}) to calculate data gradients.

An example LFM for the transcriptional regulation model from \ref{sec: transreg} is written below, demonstrating how simple \alfi{} makes constructing LFMs:

\begin{minted}[breaklines]{python}
class TranscriptionLFM(OrdinaryLFM):
    def __init__(self, num_outputs, gp_model, config):
        super().__init__(num_outputs, gp_model, config)
        self.decay_rate = Parameter(
          torch.rand((self.num_outputs, 1)))
        self.basal_rate = Parameter(
          torch.rand((self.num_outputs, 1)))
        self.sensitivity = Parameter(
          torch.rand((self.num_outputs, 1)))
        # add any constraints...

    def odefunc(self, t, h):
        f_latents = self.f
        f_latents = self.G(f_latents)  # optionally add non-linearity
        dh = self.basal_rate + self.sensitivity * f_latents - self.decay_rate * h
        return dh

\end{minted}

\subsection{Unified Form Language for PDEs}\label{app: ufl}

The \texttt{PartialLFM} requires an additional module called \texttt{fenics\_module}. This is defined in Unified Form Language (UFL) \citep{alnaes2014unified}. An example UFL snippet for the reaction-diffusion equation from Section \ref{sec: reactdiff} is shown below: 

\begin{minted}[bgcolor=lightgray,breaklines]{python}
mesh = UnitIntervalMesh(20)
V = FunctionSpace(mesh, 'P', 1)

u = TrialFunction(V)
v = TestFunction(V)

a = (1 + dt * decay) * u * v * dx 
a = a + dt * diffusion * inner(grad(u), grad(v)) * dx
L = (u_prev + dt * sensitivity * latent_force) * v * dx

bc = DirichletBC(V, 0, 'on_boundary')

y = Function(V)
solve(a == L, y, bc)
\end{minted}

\section{Datasets} \label{app: trans datasets}

For the transcriptional regulation experiments, we used two datasets: the microarray data as used in \citet{lawrence2007modelling} (accessible with ArrayExpress ID E-MEXP-549 \citep{barencodataset} as well as a larger RNA-sequencing dataset (accessible with GEO Series GSE100099 \citep{hafner2017p53}), both pertaining to the p53 regulatory network. The p53 protein is crucial in DNA damage response and is activated by ionising radiation used in radiotherapy. The first data comes from human lymphoblastic leukemia (white blood cell cancer) cells $\gamma$-irradiated with 5 Gy, 2.45 Gy/minute, and sampled every 2 hours for 12 hours. The RNA-sequencing data comes from the human breast cancer cell line MCF7. Similarly, the cells are $\gamma$-irradiated with 10 Gy and sampled every hour for 12 hours. All the real datasets used in our experiments are anonymised or do not contain personally identifiable information. Moreover, they are all publicly available.

\section{Time profiling} \label{app: time}
\begin{figure}[h]
     \centering
          \begin{subfigure}[b]{0.45\columnwidth}
         \centering
    \includegraphics[width=\textwidth]{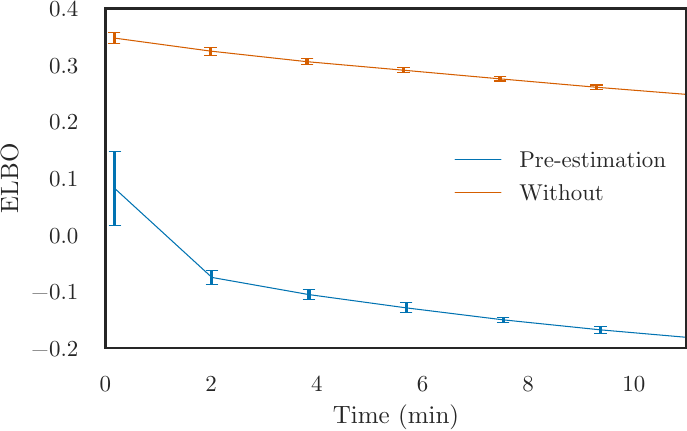}
    \caption{Latent Force Model.}
    \label{fig:lfm preest}
     \end{subfigure}
     \hfill
 \begin{subfigure}[b]{0.45\columnwidth}
         \centering
    \includegraphics[width=\textwidth]{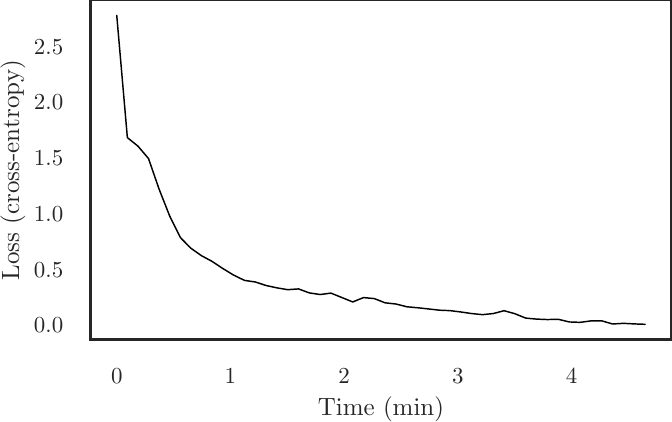}
  \caption{Latent Force Operator.}
         \label{fig:lfo loss}
     \end{subfigure}
        \caption{Training times compared between the single-instance Latent Force Model and the Latent Force Operator. Training was run on a Titan Xp GPU.}
        \label{fig:lfolfm training comparisons}
\end{figure}

\begin{figure}[h]
    \centering
    \includegraphics[width=\columnwidth]{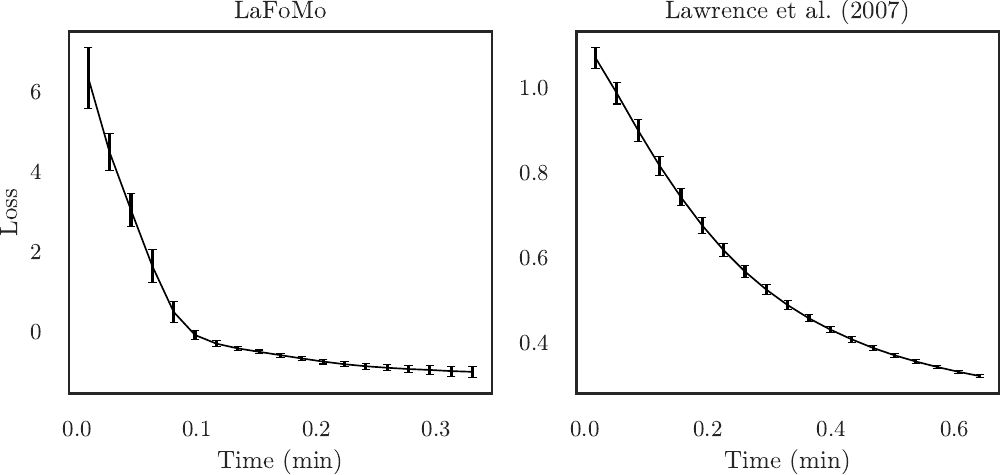}
    \caption{Convergence times of \alfi{} compared with the analytical model in \citet{lawrence2007modelling}.  We found that increasing the learning rate for the latter model lead to instabilities in the cholesky decompositions. Both were implemented with GPyTorch backends and run on an AMD Ryzen 5600X CPU.}
    \label{fig:1d convergence}
\end{figure}

As shown in Figure \ref{fig:lfolfm training comparisons}, training the Latent Force Model (using FEniCS on a single instance) takes a comparable amount of time to training a Latent Force Operator (using the Fourier Neural Operator) on a dataset of 4,000 instances. Moreover, the convergence of the variational LFM has a faster or comparable training time to the analytical solution, as shown in Figure \ref{fig:1d convergence} for the 1D case.

At prediction time, the LFO takes $(6.25 \pm 0.16)\times 10^{-4}$ minutes compared with approximately 12 minutes for convergence of the standard variational LFM, an almost $20,000\times$ speedup. 

\section{Additional Results}

\subsection{Lotka-Volterra} \label{app: lotka}
\begin{figure}[h]
    \centering
    \includegraphics[width=\columnwidth]{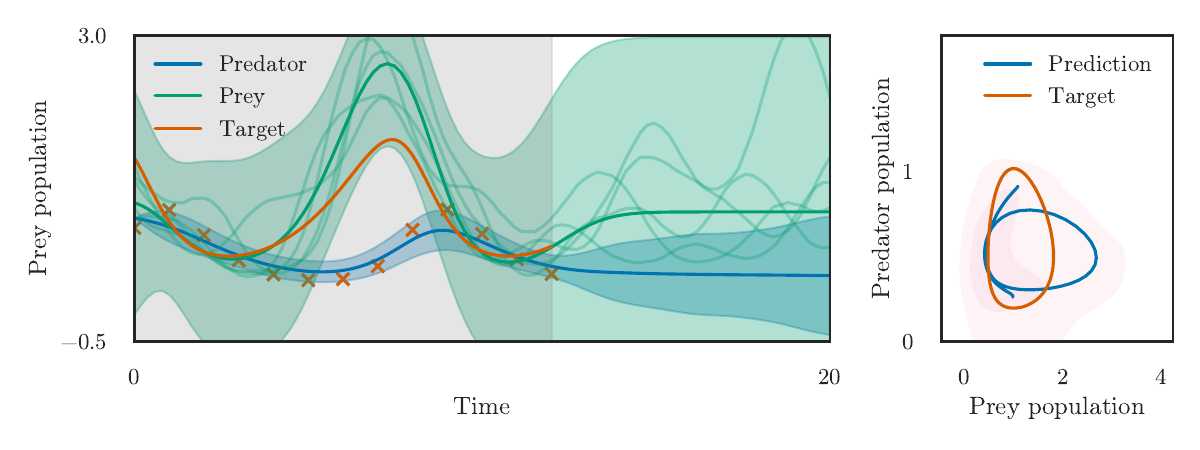}
    \caption{\alfi{} with an RBF kernel on the Lotka-Volterra task. This demonstrates inferior extrapolation performance compared with a periodic kernel, since a periodic kernel has additional inductive biases.}
    \label{fig:lotka-rbf}
\end{figure}

In Figure \ref{fig:lotka-rbf}, the results for the Lotka-Volterra task are shown when using an RBF kernel instead of the periodic kernel.

\begin{figure*}[t]
    \centering
    \includegraphics[width=0.9\textwidth]{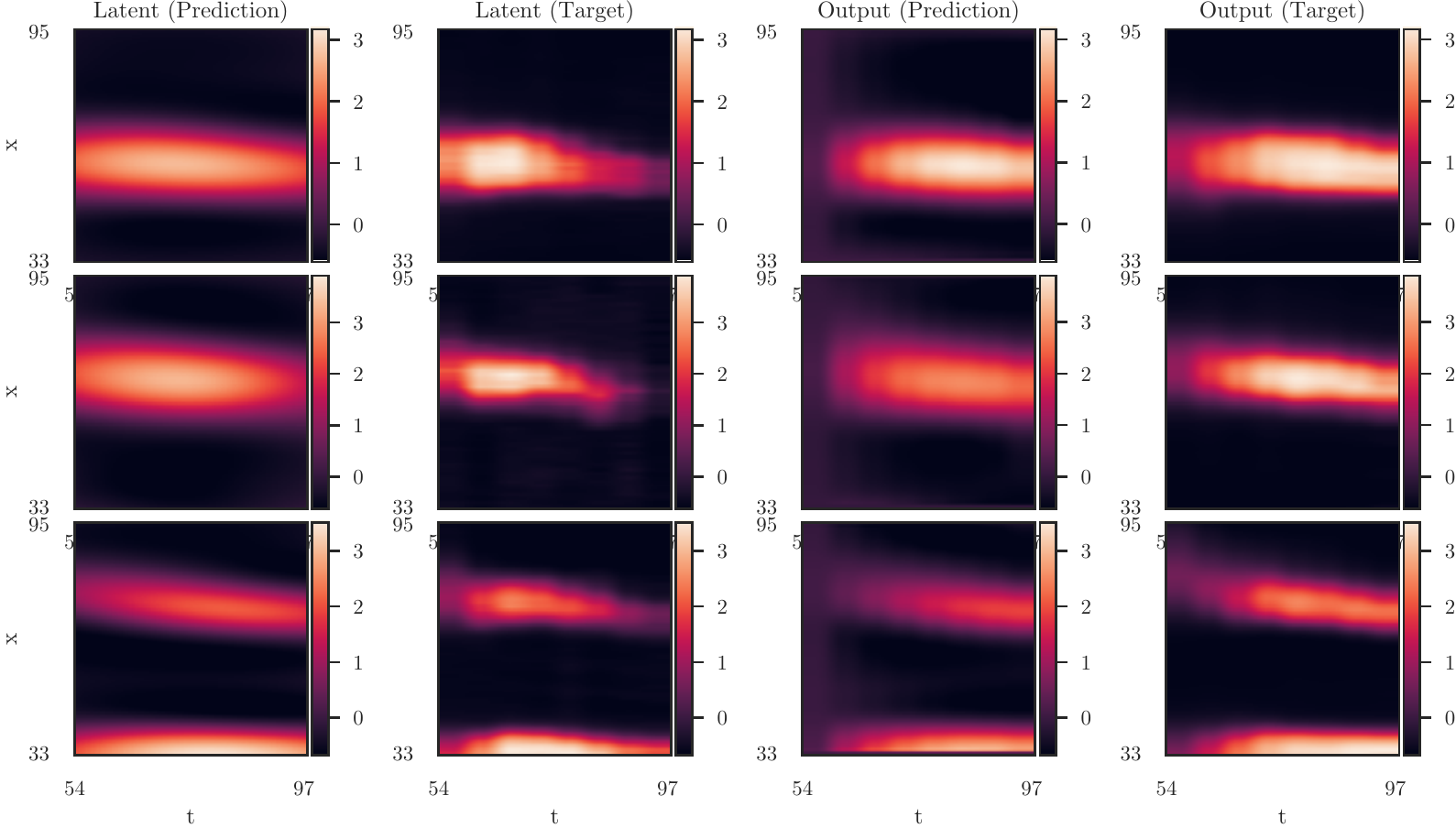}
    \caption{\alfi{} on the Drosophila reaction-diffusion task. \textit{First and second columns}: the latent mRNA concentration alongside ground truth. \textit{Third and fourth columns:} protein concentration alongside ground truth.}
    \label{fig:dros_full}
\end{figure*}

\subsection{Drosophila}  \label{app: dros}

In Figure \ref{fig:dros_full}, the results for the reaction-diffusion experiment with the Drosophila embryogenisis dataset are shown.

\section{Hyperparameters}\label{app: hyper}

Hyperparameter optimisation is typically carried out with maximum marginal likelihood, or a lower-bound thereof. The hyperparameters were initial set to reasonable values: lengthscales of kernels were set to the smallest distance between datapoints, and rates to a uniform random number between 0 and 1. For the comparison with GP-mRNA \citep{lopez2019physically}, we kept the experimental conditions the same: the rates were fixed to the values reported by \citet{becker2013reverse}, and the lengthscales were learnt by maximising the ELBO.

\newpage
\hspace{1cm}
\newpage
\bibliographystyle{aaai}
\bibliography{refs}
\end{document}